\documentclass{article}
\usepackage[utf8]{inputenc}
\usepackage{amssymb}
\usepackage{amsmath}
\usepackage{multirow}
\usepackage{chngcntr}
\usepackage{xcolor}
\usepackage{hyperref}
\usepackage{graphicx}
\usepackage{booktabs}
\usepackage[margin=1in]{geometry}
\usepackage{pifont}
\newcommand{\cmark}{\ding{51}}
\newcommand{\xmark}{\ding{55}}

\counterwithin{table}{section}

\title{Layered gradient accumulation and modular pipeline parallelism: fast and efficient training of large language models}
\author{Joel Lamy-Poirier\thanks{ServiceNow, joel.lamy-poirier@servicenow.com}}

\begin{document}

\maketitle

\begin{abstract}

The advent of the transformer has sparked a quick growth in the size of language models, far outpacing hardware improvements. (Dense) transformers are expected to reach the trillion-parameter scale in the near future, for which training requires thousands or even tens of thousands of GPUs. We investigate the challenges of training at this scale and beyond on commercially available hardware. In particular, we analyse the shortest possible training time for different configurations of distributed training, leveraging empirical scaling laws for language models to estimate the optimal (critical) batch size. Contrary to popular belief, we find no evidence for a memory wall, and instead argue that the real limitation --- other than the cost --- lies in the training duration. 

In addition to this analysis, we introduce two new methods, \textit{layered gradient accumulation} and \textit{modular pipeline parallelism}, which together cut the shortest training time by half. The methods also reduce data movement, lowering the network requirement to a point where a fast InfiniBand connection is not necessary. This increased network efficiency also improve on the methods introduced with the ZeRO optimizer, reducing the memory usage to a tiny fraction of the available GPU memory.

\end{abstract}

\section{Introduction}
\label{sec:intro}

Large scale language models are rapidly changing the field of natural language processing. Transformers \cite{vaswani2017attention} have risen as the preferred architecture for language models, being simpler, more scalable and more performant than alternatives based on recurrent neural networks. Although transformers come in numerous flavors and variations \cite[\dots]{brown2020language,devlin2019bert,fedus2021switch,lewis2019bart,radford2019language,raffel2020exploring,yang2020xlnet}, a growing body of work suggests that the model performance is mainly driven by scale. Furthermore, few-shot learning capabilities have been observed in GPT-3 (175 billion parameters) \cite{brown2020language}. This opens the field of natural language processing to a wide range of new applications where fine-tuning is either impossible or impractical. These few-shot learning capabilities have not been observed in smaller models, so they are believed to emerge only at the scale of GPT-3, reinforcing the need for extremely large language models.

While large language models present exciting new capabilities, they also pose a difficult and costly engineering challenge. For example, simply storing the training state of GPT-3 takes about 2 terabytes of memory, and storing its intermediate activations and gradients takes several more terabytes. This is dozens of times more memory than available on the largest GPU to date, the 80 GB NVIDIA A100. Besides memory, training also requires a gigantic amount of computing power. GPT-3 needs about 3600 petaflop-day to train, or 30 years on the same device.

To speed up training of large models, it is necessary to parallelize the training process by distributing the load across multiple GPUs. The leading method is 3d parallelism (see for example \cite{shoeybi2020megatronlm}), which combines the three common form of parallelism: data, pipeline and tensor parallelism. Together with methods such as mixed precision training and activation checkpointing, 3d parallelism allows quickly training large models with up hundreds of billions of parameters but is not so fast beyond that scale. For example, GPT-3 was trained in a matter of days, while the trillion-parameter version of Megatron-LM would need more than three months to train \cite{narayanan2021efficient}. For larger models, training times are in the order of years or worse. 

Recently, several memory optimizations have been suggested for large models, aiming in particular to simplify fine-tuning. When compared to training from scratch, fine-tuning requires much less computational power and can be done with a limited number of GPUs, but doing so creates a memory bottleneck. The ZeRO family of methods addresses this bottleneck by partitioning the training state \cite{rajbhandari2020zero}, aggressively offloading memory \cite{ren2021zerooffload,rajbhandari2021zeroinfinity}, and breaking down individual operations \cite{rajbhandari2021zeroinfinity}. These methods together shatter any memory constraint.

While a lot of attention has been given to the memory usage, much less effort has been dedicated to reducing the training time. To that end, we investigate various parallel training configurations and strategies for large and dense transformers, with the goal of minimizing the training time on existing hardware. We explicitly integrate the concept of \textit{critical batch size} \cite{golmant2018computational,mccandlish2018empirical,shallue2019measuring} in our analysis, leveraging the empirical scaling laws found in \cite{kaplan2020scaling}. The critical batch size provides an upper bound on the efficient scaling of the batch size, and by extension dictates how many GPUs can be used for training. To our knowledge, our analysis is the first to directly integrate the critical batch size and the resulting parallelism bounds for large language models. Our analysis can be applied to a wide range of scales, from the tiniest thousand-parameter transformers up to the quadrillion parameter scale and beyond.

Our analysis shows that memory is not a limiting factor even past the trillion-parameter scale. Instead, we find a computational bottleneck caused by the limitations of distributed training. Due to constraints from the critical batch size and network connectivity, 3d parallelism can efficiently use a limited number of GPUs, and this upper bound does not increase nearly as fast as the amount of computation needed to train larger models. We find a minimum training time of about two weeks for a trillion-parameter model, and this bound scales worse than linearly with respect to the model size. As a result, training times are in the order of months or years above the trillion-parameter scale. We also find that the ZeRO family of methods counter-productive for the training time, largely because of frequent data transfers in micro-batched approaches --- including pipeline parallelism --- which prevent efficient 3d parallelism.

We introduce two closely related methods which significantly improve the training efficiency, while also reducing the memory usage and network requirement. The first one, \textit{layered gradient accumulation}, uses a bandwidth-efficient scheduling for gradient accumulation, which makes it easier to overlap the gradient reduction with the computation. It also reconciles gradient accumulation with the methods introduced in the ZeRO optimizer, avoiding frequent data transfers when partitioning the training state or when offloading it to CPU memory.The second method, \textit{modular pipeline parallelism}, uses a modular split of the layers to improve the efficiency of pipeline-parallel training by minimizing the pipeline ``bubble'', which otherwise limits the efficiency of pipeline parallelism. It also builds upon layered gradient accumulation, enabling its benefits in the pipeline-parallel case. In particular, it allows partitioning the training state in the fastest 3d parallel settings, reducing the memory usage to a minimum and preventing a trade-off between memory usage and the bubble reduction of modular pipeline parallelism. These methods together allow training at least twice as fast as previous methods, for example reducing the minimum training time to one week for a trillion-parameter model. While we focus on large transformers, layered gradient accumulation and modular pipeline parallelism are not specific to transformers or even large models. In fact, the improved communication overlap is particularly useful for smaller models, especially over slower networks. 

Concurrent to our work, two papers appeared that show overlap with our results. The latest version of Megatron-LM \cite{narayanan2021efficient} suggests a breakdown of the layers similar to modular pipeline parallelism, however the authors suggest a different scheduling aimed at reducing the activation memory. Our method instead focuses on a network-efficient method, which leads to an increased efficiency benefit from the new layer breakdown and when combined with a training state partition also leads to a lower memory usage. In section \ref{sec:fast_checkpoints}, we investigate disk offloading in a way similar to ZeRO-Infinity \cite{rajbhandari2021zeroinfinity}. Our methods show improved results, further reducing the requirements for offloading and enabling offload even on slow hard drives. However, we also little use for this extra space, as the memory usage tends to remain reasonable in most scenarios. Instead, we suggest leveraging the results to improve checkpointing methods.

This paper is structured as follow. In section \ref{sec:overview} we describe the existing approaches and provide some additional context for the paper. In sections \ref{sec:layer_grad_accumulation} and\ref{sec:modular_pipeline} we introduce the main new methods. In section \ref{sec:methodology}, \ref{sec:trillion_parameter} and \ref{sec:scaling_analysis}, we analyze the impact of our method on training speed and memory usage, then generalize to a wide range of scales. In section \ref{sec:additional_considerations} we investigate some additional concerns that arise in realistic training scenarios. Finally, in section \ref{sec:conclusion} we discuss the implications and limitations of our results.

\section{Background and related work}
\label{sec:overview}

In this section we describe the optimizations and distributed training methods relevant to our analysis.

\subsection{Critical batch size}
\label{sec:critical_batch}

In stochastic gradient descent, the gradient of the loss with respect to the model parameters is estimated from a micro-batch. When the micro-batch is small, adding more samples improves this estimate and allow training with a proportionally lower number of steps, keeping the total amount of computation unchanged. However, past a certain point, the estimate becomes accurate enough and adding more samples no longer reduces the required number of steps. The \textit{critical batch size} \cite{golmant2018computational,mccandlish2018empirical,shallue2019measuring} provides an upper limit for the small micro-batch regime, and can obtained by estimating the variation (noise) of the gradients between individual samples, relative to the true gradients \cite{mccandlish2018empirical}.

\subsection{Mixed precision training}
\label{sec:mixed_precision}

Deep neural networks do not require a high numerical accuracy, and to some extent can be trained with as little as 16 bits of precision. The state-of-the-art approach in that regard is \textit{mixed precision training} \cite{micikevicius2018mixed}, in which the bulk of the computation is done in half-precision, while the weights are stored and updated in single-precision. Half-precision however has a limited exponent range, and the gradients need to be dynamically scaled to limit the risk of underflows and overflows. The problem can also be solved with the \textit{bfloat16} data format, which has a wider exponent range and is available in recent devices such as the NVIDIA A100.

\subsection{Communication overlap}

In addition to the computation itself, data transfers present a significant optimization challenge. They should preferably be done in parallel with the bulk of the computation, otherwise the computational cores stay idle during the transfer. As the computation and network communications mostly use separate resources, they can in principle be \textit{overlapped} with near-perfect efficiency, so the runtime is determined by the slowest of the operations. The operation is said to be \textit{compute-bound} or \textit{data-bound} depending on which one finishes last, with the former scenario being preferable here. In practice, there is a small overhead to overlapping the operations, but we ignore it for the purpose of this paper.

The threshold for a compute-bound operation is described through the concept of \textit{arithmetic intensity}. For an operation which requires both computation and data transfer, the arithmetic intensity is defined as the ratio between the amount of computation and data transfer. In the case of perfect overlap, the operation is compute-bound if the arithmetic intensity is higher than what the hardware can support. For example, A NVIDIA A100 has a peak computational power of 312 teraflops for half-precision and a memory bandwidth of 2039 GB/s, for an arithmetic intensity threshold of 143 flops/B. A binary addition with an arithmetic intensity of 1/6 lies deeply in the memory-bound region, while the multiplication of two 1024x1024 matrices has an arithmetic intensity of 341 and is compute-bound.

\subsection{Distributed training}
\label{sec:distributed}

There are two main forms of parallelism, data parallelism in which each device performs the same computation on a subset of the data, and model parallelism in which the model and computation are split between devices.


In \textbf{data parallelism}, the input batch is split between the $n_b$ devices. Each device independently processes its assigned input, then the resulting gradients are \textit{reduced} (summed) between the devices, and finally each independently updates all the weights. All the network communication happens in the gradient reduction, which for the most part can be overlapped with the backward pass, and this overlap can be made compute-bound with large enough micro-batches. Data parallelism is the most used method for parallel training, because of its simplicity and wide availability. However, as each device needs to store a copy of the model, the memory usage is excessive for larger models. This can be addressed by partitioning the model, as suggested in \cite{rajbhandari2020zero}. In this partitioned case, the model is split between the devices, each being responsible for storing and updating an equal share of the weights\footnote{In this paper we consider a partition of the whole training state, which in the terminology of \cite{rajbhandari2020zero} corresponds to the third stage of ZeRO-DP.}. The weight tensors are reconstructed on each device as needed, and the reduced gradients are kept only as required. The partition increases the network communication by 50\%. Data parallelism scales up to the critical batch size $b_c$, but in general a minimum micro-batch size is required for efficient communication overlap, reduces the limit to a fraction of $b_c$.

\begin{figure}[t]
\centerline{\includegraphics[scale=1, trim=0 20pt 20pt 20pt]{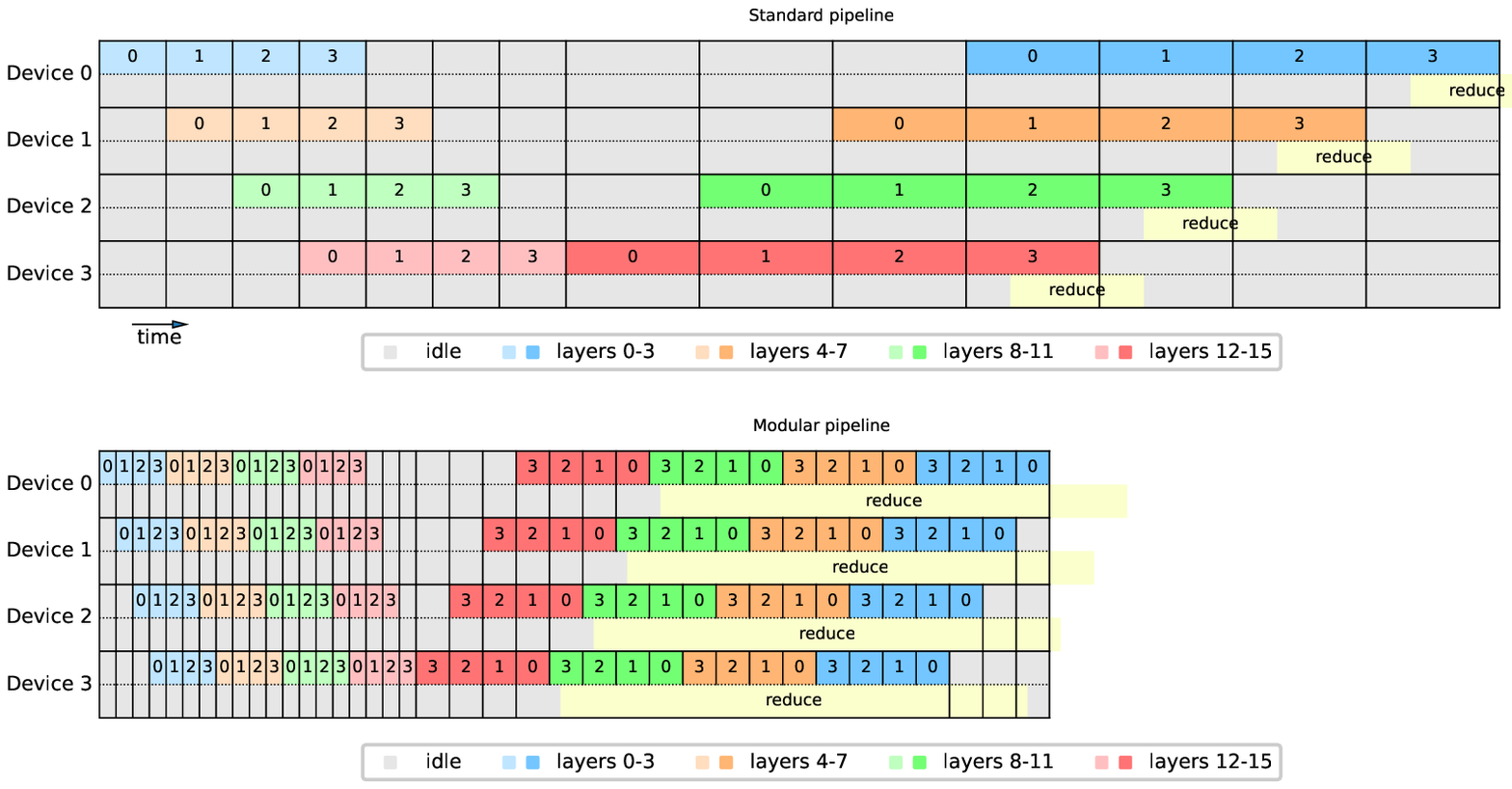}}
\caption{Computation and network scheduling example with standard and modular pipelines (no state partition). The modular version significantly reduces the idle time from the bubbling effect and makes the gradient reduction much easier to overlap.}
\label{fig:pipeline_comparison}
\end{figure}

Model parallelism comes in two main forms, depending on how the model is split. The first form is \textbf{pipeline parallelism}, where each of the $n_l$ devices is responsible for a subset of the layers. The input is pipelined through the devices, and parallel computation is achieved by feeding micro-batches sequentially to the model \cite{huang2019gpipe}. Each device processes all the micro-batches for a given batch, then updates the weights. This however leads to a ``bubbling'' effect in which some of the devices stay idle due to input starving (see the upper part of figure \ref{fig:pipeline_comparison})\footnote{The network bubble can be avoided with \textit{asynchronous training} \cite{niu2011hogwild,narayanan2021memoryefficient}, where the weight updates are interleaved with the computation. However, this leads to gradients being computed on \textit{stale} (outdated) weights, which can be harmful to training. Although training is technically done with sequential batches rather than micro-batches, \textit{staleness} adds to the gradient noise and reduces the critical batch size proportionally to the delay \cite{duchi2015asynchronous,hannah2017unbounded,stich2021critical}. Consequently, the limitations of 3d parallelism (see below) are the same as with \textit{synchronous training}.}. For an input split into $n_\mu$ micro-batches, this bubble increases the training time by a factor $\tfrac{n_l-1}{n_\mu}$. Pipeline parallelism is also limited in magnitude, as it cannot grow beyond the number of layers in the model.

The other form of model parallelism, \textbf{tensor parallelism}, involves slicing tensors so that each device is responsible for evaluating a slice of the activations using its corresponding slice of the weights\footnote{The method is often referred to as ``model parallelism'' in the literature, but the term may also refer to pipeline parallelism, so here \textit{tensor parallelism} is used instead, and the more general term ``model parallelism'' is reserved for the combination of both methods.}. Tensor parallelism is more complex than the two other forms of parallelism and generally depends on the internal details of the model. It also requires a lot more network communication and is in general only feasible with the fastest interconnects such as NVLink. In practice this limits tensor parallelism to 16 GPUs, the maximum that can be connected with NVLink.

Although data, pipeline and tensor parallelism are individually limited, they can be combined together into (up to) \textbf{3d parallelism}. The dimensions combine in a multiplicative way, for a total of  $n_\text{gpu}=n_b n_l n_a$ devices. In 3d parallelism, the input batch size is split both between the data-parallel instances\footnote{The term \text{instance} is used to designate a slice of the cluster in a specific parallelism dimension. For example, a data parallel instance refers to the $n_l n_a$ devices handling the same sequence of micro-batches.}, and between the $n_{\mu}\geq n_l$ sequential micro-batches. For a batch of size $b$, this implies a micro-batch size $b_{\mu}=b/(n_b n_{\mu})$. Because of the limitation from the critical batch size (see section \ref{eq:critical_batch}), this implies a competition between data and pipeline parallelism, but the combined method generally allows a higher degree of parallelism than with either method in isolation. This is because pipeline parallelism increases the arithmetic intensity for the gradient reduction, reducing the minimum micro-batch size\footnote{\textit{Gradient compression} methods such as \textit{1-bit Adam} \cite{tang20211bit} also reduce the network communication, potentially providing the same benefit without pipeline parallelism. However, compression methods cannot efficiently be used with a partitioned training state. Together with the absence of pipeline parallelism, this makes the size of the training state difficult to manage for larger models.}. However, pipeline parallelism provides no such benefit with a partitioned training state, where the network operations need to be repeated for each micro-batch.

\subsection{Memory optimizations}
\label{sec:memory_optimizations}

As the model grows in size, so does its memory usage, and substantial memory optimizations are needed to avoid running out of memory. The bulk of the memory usage falls in two categories: the training state and the activation memory. The \textit{training state} consists of the model parameters and the optimizer state, which for Adam includes the running average and variance of the parameters. For the purpose of this paper, the parameter gradients are also included in the training state. The \textit{activation memory} consists of the layer activations and their gradients. Both categories can use substantial amounts of memory for larger models, but various techniques have been developed to reduce memory consumption.

The training state is proportional to the model size. At 80 GB, a NVIDIA A100 can store at most 20 billion parameters in single precision, or 6.7 billions with ADAM. As the training size is fixed, additional memory is needed to fit larger models, and distributed training provides such memory. As described in section \ref{sec:distributed}, the training state is split by construction in the model-parallel directions and can be partitioned in the data parallel direction. The training state can also be offloaded to CPU memory; however, this requires additional data transfers and may create a communication bottleneck.

With a training state partition or offloading, the model weights are stored in another device and must be restored to the before usage, so additional buffers are required. Similarly, the gradients need to be created on a buffer before being moved to the appropriate place. While these \textit{parameter and gradient buffers} are small compared to the whole training state, they dominate the GPU memory usage for larger models once all the optimizations are taken into account. These buffers are to some extent also required in the absence of partition or offloading, as for example the parameters need to be converted to 16 bits before usage.

Activation memory is also a concern, and naive implementations can easily require hundreds of gigabytes even at the billion-parameter scale. The most significant reduction in activation memory is obtained with \textit{activation checkpointing}, in which the bulk of the activations are dropped during the forward pass \cite{chen2016training}\footnote{The method is also known as \textit{gradient checkpointing} in the literature, even though it does not involve any gradient.}. A subset of the activations is kept as \textit{checkpoints} and are used to recalculate the remaining activations during the backward pass. This lowers the activation memory by a significant factor, reducing it to the \textit{layer activations}, i.e., the activation memory required between the activation checkpoint, and the \textit{activation checkpoints} themselves, which can in many cases be offloaded to CPU memory. The method comes at the cost of a 33\% increase in computation, but there is no alternative for larger models. 

As the activation memory is proportional to the micro-batch size, an obvious memory optimization is to lower it, down to a single sample if possible. While single-sample micro-batches are typically not recommended, they can run efficiently for larger models for which the computation kernels are big enough. However small micro-batches come with a lower arithmetic intensity with respect to the model weights, potentially creating a bottleneck in the case of data parallelism or memory offload. The \textit{layered gradient accumulation} and \textit{modular pipeline parallelism} methods introduced in this paper are designed to prevent such bottlenecks, and to allow running efficiently with a micro-batch size of one.

On a side note, \textit{gradient accumulation} allows increasing the batch size without running out of memory, by processing multiple micro-batches sequentially between weight updates. This is especially useful with data parallelism over a slow network, as larger batches reduce the frequency of the gradient reduction. However, this leads to inefficient communication overlap as the gradient reduction is concentrated in the last micro-batch. The method is also counter-productive for optimizing the training time, since the micro-batches are processed sequentially instead of in parallel. The \textit{layered gradient accumulation} is designed to improve the communication overlap, while pipeline parallelism allows processing the micro-batches in parallel.

In addition to the training state and activation memory, there can be some additional memory usage or hidden memory costs. Many libraries for distributed training, including Pytorch \textit{DistributedDataParallel}, 
use a bucketing strategy for network operations, combining several tensors in a temporary \textit{network buffer}. This is preferable for small tensors, as it avoids the overhead or running the individual operations sequentially, but comes at the cost of additional memory usage and data movement. In ZeRO-R it is suggested to use a constant size buffer to keep the memory overhead bounded~\cite{rajbhandari2020zero}. However, in the present case, the parameter and gradient buffers already play a role similar to the buckets, and \textit{in-place network operations} can be done at no extra memory cost.

Even when enough memory is available, \textit{memory fragmentation} can still cause the device to run out of memory. For larger models, memory fragmentation becomes a problem because the allocations involve very large contiguous chunks of memory, and there is an overlap between short-lived activations and their gradients, and longer-lived tensors such as the activation checkpoints and parameter gradients. A solution to this is to pre-allocate contiguous buffers for the memory whenever possible. Pre-allocation also allows combining tensors into a single contiguous (fused) chunk of memory, which not only ensures optimal allocation but also allows running data operations on a fused tensor. In~\cite{rajbhandari2020zero} it is suggested to pre-allocate the activation checkpoints and parameter gradient.
This leaves the layer activations and gradients as the main source of fragmentation, which can still be significant because of the size imbalance between the various activations, which creates large memory gaps. For example, we observed an overhead of up to $40\%$ for a single transformer layer in the PyTorch implementation.
However, it is possible to reduce this overhead to almost zero with a memory-efficient implementation, so in this paper we assume the memory fragmentation cost to be minimal.

\section{Layered gradient accumulation}
\label{sec:layer_grad_accumulation}

In \textit{layered gradient accumulation}, we split the input into micro-batches exactly as in standard gradient accumulation, but we process all the micro-batches for a given layer before proceeding to the next one. We take such layers as the intervals between activation checkpoints, so that we can drop the intermediate activations between the micro-batches.

\begin{figure}[t]
\centerline{\includegraphics[scale=0.98, trim=0 0 20pt 0]{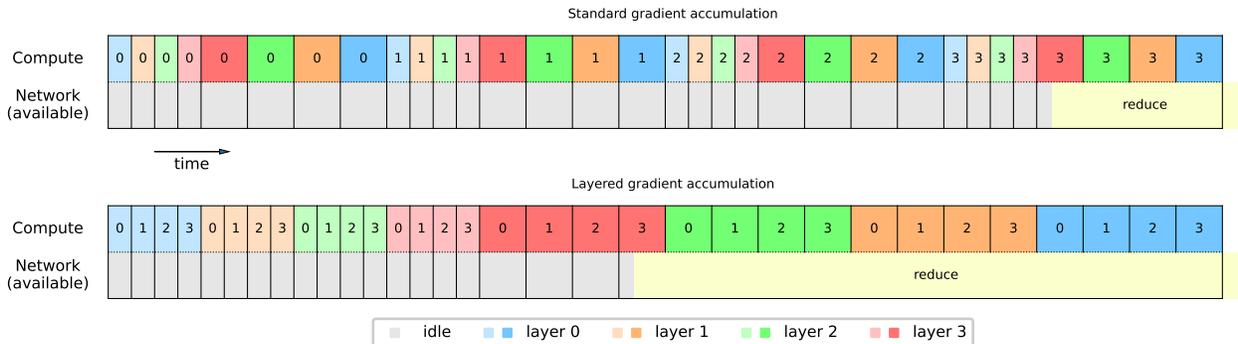}}
\caption{Computation and network scheduling example with data parallelism for standard gradient accumulation (top) and layered gradient accumulation (bottom). The colors represent the different layers, with different shades for the forward and backward pass (lengths not to scale), and the numbers indicate the micro-batch index. The layered version reduces the network requirement by spreading the gradient reduction over most of the backward pass.}
\label{fig:gradient_accumulation}
\end{figure}

\begin{figure}[t]
\centerline{\includegraphics[scale=1, trim=0 0 25pt 0]{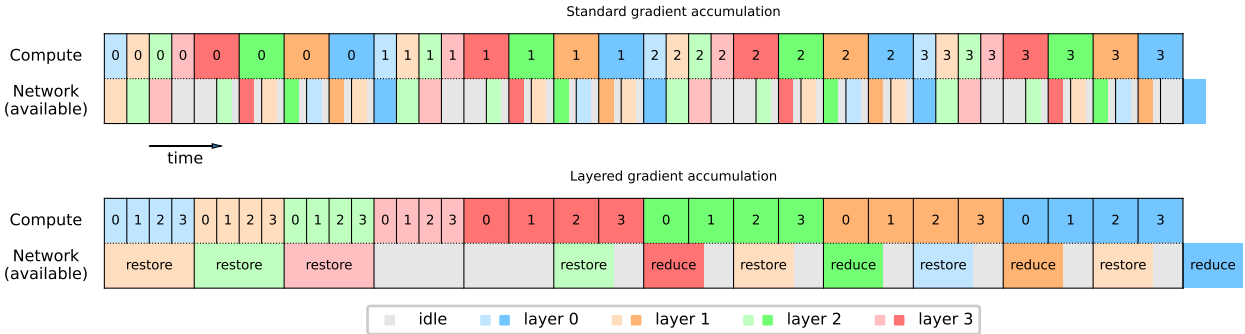}}
\caption{Computation and network scheduling example with state partition or offload, for standard gradient accumulation (top) and layered gradient accumulation (bottom). A mixed buffering method is assumed (see appendix \ref{sec:parameter_buffer_details}). The standard version involves frequent context switches with respect to the weights, which leads to unreasonable bandwidth requirements. The layered version on the other hand maximizes the reuse of the restored weights, so requires the same bandwidth as without gradient accumulation.}
\label{fig:gradient_accumulation_partitioned}
\end{figure}

The layered gradient accumulation method is advantageous from a data perspective. With data parallelism, it allows an efficient overlap of the gradient reduction with the backward pass, unlike traditional gradient accumulation which only allows overlapping with the last micro-batch. This is illustrated in figure \ref{fig:gradient_accumulation}. With a state partition, layered gradient accumulation greatly reduces the bandwidth requirement by eliminating redundancies in the parameter restoration and gradient reduction operations, as illustrated in figure \ref{fig:gradient_accumulation_partitioned}. Similarly, layered gradient accumulation helps with offloading by reducing the amount of data movement. All the activation checkpoints must be kept, which in the presence of pipeline parallelism is already a requirement\footnote{
Some approaches perform better with a memory-efficient scheduling \cite{narayanan2021memoryefficient,narayanan2021efficient}, but in any case the activation checkpoints do not grow too big.}, but otherwise may cause an increase in the activation checkpoint memory.

Note that layered gradient accumulation generally cannot be efficiently combined with standard pipeline parallelism. Indeed, a given pipeline-parallel instance must process every micro-batch for all every layer other than the last before it can pass an output to the next instance. This is addressed with the modular pipeline parallelism.

\section{Modular pipeline parallelism}
\label{sec:modular_pipeline}

In pipeline parallelism, the layers are generally split into contiguous chunks. For a network with $d_l$ layers split into $n_l$ devices, first instance gets the layers 1 to $d_l/n_l$, the second gets the layers $d_l/n_l+1$ to $2d_l/n_l$, etc. However, while this ``naive'' splitting minimizes pipeline-parallel network operations, it also maximizes the bubbling effect. In modular pipeline parallelism, the layers are instead split in a modular fashion, so that the first instance gets the layers 1, $n_l+1$, etc., the second gets the layers 2, $n_l+2$, etc., and so on. The computation is scheduled as with layered gradient accumulation, i.e., a given instance processes all micro-batches for a given layer, then goes on to the next layer for which the input should be ready, etc.

In the modular formulation, a micro-batch reaches the last instance after being processed on $n_l-1$ layers rather than $d_l(1-1/n_l)$, reducing the bubbling overhead by a factor $d_l/n_l$. This makes it possible to reduce the bubbling to almost zero without increasing the number of micro-batches $n_\mu$. Additionally, the gradient reduction is spread more evenly over the backward pass, dividing the network overhead by a factor $d_l/n_l$.

Modular pipeline parallelism comes with an increased pipeline parallel network cost since data needs to be transferred after each layer, but for large models this cost remains far below the data parallel network usage. This data transfer can be overlapped with computation provided there are slightly more micro-batches than pipeline-parallel instances.

\section{Methodology}
\label{sec:methodology}

In the following sections we analyse the resource usage and training time for large language models, focusing on the impact of the methods introduced in this paper, as well as 3d parallelism and training state partitioning. We present some example and relevant results, leaving the detailed computation to the appendix \ref{sec:resource_usage}.  We assume the computation is done with a scalable cluster with up to 16 A100 GPUs per node, supporting both InfiniBand and NVLink. See appendix \ref{sec:hardware} for additional details on the hardware.

\paragraph{Model}

We consider a transformer encoder following the original approach \cite{vaswani2017attention}, up to computationally unimportant modifications. Transformer encoders are for example used in the \textit{BERT} model \cite{devlin2019bert} and its derivatives, and our results generalize straightforwardly to decoder-based models such as the GPT family \cite{radford2018improving,radford2019language,brown2020language}. For simplicity, we restrict to the transformer part of the model, and ignore other components such as the tokenizer, the embedding layer and the language model head.

A transformer encoder consists of $d_l$ identical layers, each composed of a multi-head attention module followed by a non-linearity. The former consists of $d_a$ attention heads of size $d_h$, for a layer width $d_m=d_a\times d_h$, while the latter consists of a two-layer dense feedforward network with intermediate size $d_I=n_I\times d_m$. Each layer holds $p_l \approx (4+2n_I)d_m^2$ parameters, for a total of $p \approx (4+2n_I)d_m^2 \times d_l$ parameters.

For concreteness, we consider the  X$_{[x]}$ family of models described in appendix \ref{sec:transformer_scaling}, which allows extrapolation to wide range of scales. The exact configuration however makes little difference, so our results straightforwardly generalize to most configurations.

\paragraph{Training}
We use mixed precision training, and activation checkpoints at the output of each transformer layer. We overlap communication when possible and offload the training state and activation checkpoints when needed (and possible). We use the \textit{mixed gradient buffering} method for the parameter and gradient buffers, as described in appendix \ref{sec:parameter_buffer_details}. This method is not necessarily the most memory efficient but combines well with the bandwidth-efficient methods introduced here and is sufficient for all practical purposes. The Adam optimizer is assumed.

We consider distributed training with up to three parallelism dimensions. For tensor parallelism, we follow the transformer-specific method proposed in \cite{shoeybi2020megatronlm}. This approach minimizes the network communication, which however cannot be overlapped with computation.

We investigate three training strategies. In the \textit{baseline}, we use standard data and pipeline parallelism (without partition), as described in section \ref{sec:distributed}. In the \textit{partitioned approach}, we also partition the training state in the data-parallel direction. In the \textit{improved} approach, we implement layered gradient accumulation and modular pipeline parallelism. Unless specified otherwise, we partition the training state in the improved approach, as it is preferable to do so in most cases.

\paragraph{Optimal configuration}

We select the distributed training configuration as follow, with the goal of optimizing the training speed without wasting too much computational power. We train at or slightly below the critical batch size, as training above it is inefficient. The selection is based on the resource usage described in appendix \ref{sec:resource_usage}.

In the baseline, we select the highest possible pipeline parallelism (if applicable), as it reduces memory usage while also increasing efficiency due to the faster gradient reduction. We select a micro-batch count slightly above $n_l$ to ensure overlap of the pipeline-parallel computation and the smallest micro-batch size that does not cause a communication bottleneck. With pipeline parallelism, we impose a maximum overhead of 25\% from the gradient reduction, although it is only constraining for smaller models and slow networks. When offloading is needed, the micro-batch size is also constrained by the CPU-GPU transfer rate, and there may be an additional bottleneck in the PCI-express conection which is shared between CPU and Infiniband transfers. 

In the partitioned approach, we do not consider pipeline parallelism as it leads to worse results. Instead, we maximize the data parallelism degree $n_b$. Due to the increased network communication, the gradient reduction is more constraining on the micro-batch size than in the baseline approach, but offloading is in general not constraining at all due to the small size of the offloaded state.

In the improved approach, we can set the micro-batch size to one, and avoid a communication bottleneck with a high enough count $n_\mu$. We maximize $n_b$, then $n_l$ under these considerations, so that the bubbling overhead is minimized. It is in general preferable not to overlap the pipeline-parallel communication, as $n_\mu$ and $n_l$ are both small and rounding up to a single extra micro-batch would significantly reduce the training speed.

In all cases, tensor parallelism allows a near-perfect split of both the memory and computation. We impose a maximum overhead of 25\%, which for large models (above $\sim 50$ billion parameters) allows the practical limit $n_a=16$. At extreme scales (~25 trillion parameters) it becomes possible to efficiently use tensor parallelism over InfiniBand with $n_a>16$. 

\paragraph{Resource usage}

The computation, memory and bandwidth requirements are evaluated in appendix \ref{sec:resource_usage}. To obtain a training time estimate, we compare the available and required computational power, taking into account the measurable overheads from the pipeline bubble and non-overlapped data transfers. We do not however consider the efficiency of the computational kernels, the data transfers and the communication overlap, so the training times are likely underestimated. On the other hand, the memory usage is expected to be accurate for the given training configurations.

\section{Trillion parameter example}
\label{sec:trillion_parameter}

To investigate the requirements for training at the trillion-parameter scale, we consider the 1.26 trillion parameter model X$_{160}$. This model consists of 160 transformer layers with 80 heads of size 320, for a width of 25600. The sequence length is 2560, and the critical batch size is estimated to be around 2420. Training for 100 k steps requires $6.24\times 10^{24}$ floating point operations, or 72 exaflop/s$\cdot$day, which corresponds to 231 k GPU-days on A100s at perfect efficiency.

Although the computational requirement strongly hints that a large cluster is necessary, for comparison purpose we investigate various distributed training scenarios, with the training configuration selected as follow. The resulting configurations are summarized in table \ref{table:parallel_methods_comparison}, together with the expected computational efficiency and training time. We find both data and tensor parallelism to be necessary (and together sufficient) to train in a reasonable amount of time. However, modular pipeline parallelism stands out with both a high GPU count and near-optimal efficiency, allowing to train the model in a week. It also out-performs the baseline in the absence of tensor parallelism, but the training time remains above three months.

\begin{table}[t]

\caption{Fastest training configuration for X$_{160}$ for selected training methods}
\label{table:parallel_methods_comparison}
\centering
\small
\begin{tabular}{cccccccccccc} 
\toprule
   Parallelism &       Method & Offload &   $b$ & $b_\mu$ & $n_\mu$ & $n_\text{gpu}$ & $n_b$ & $n_l$ & $n_a$ & Efficiency &   Time \\
\midrule
          None &     Baseline &  \cmark &  2416 &       4 &     604 &              1 &     1 &     1 &     1 &       1.00 &  630 y \\
          Data &     Baseline &  \cmark &  2415 &       5 &       1 &            483 &   483 &     1 &     1 &       1.00 &  1.3 y \\
          Data &  Partitioned &  \cmark &  2415 &       5 &       1 &            483 &   483 &     1 &     1 &       1.00 &  1.3 y \\
   Data + pipe &     Baseline &  \cmark &  2412 &       4 &     201 &            480 &     3 &   160 &     1 &       0.56 &  2.4 y \\
   Data + pipe &     Improved &  \xmark &  2415 &       1 &       5 &           2415 &   483 &     5 &     1 &       0.94 &  100 d \\
    \hline
 Data + tensor &     Baseline &  \cmark &  2415 &       5 &       1 &           7728 &   483 &     1 &    16 &       0.93 &   32 d \\
 Data + tensor &  Partitioned &  \xmark &  2415 &       5 &       1 &           7728 &   483 &     1 &    16 &       0.93 &   32 d \\
            3d &  Baseline &  \xmark &  2408 &       1 &     172 &          \textbf{35840} &    14 &   160 &    16 &       0.48 &   13 d \\
            3d &  Improved &  \xmark &  2415 &       1 &       5 &          \textbf{38640} &   483 &     5 &    16 &       0.88 &  \textbf{6.8 d} \\
\bottomrule
\end{tabular}

\caption{Memory usage breakdown for the same training configurations.
}
\label{table:parallel_methods_memory}
\centering
\small
\begin{tabular}{cccccccc}
\toprule
   Parallelism &       Method &   State & Checkpoint & Buffers & Activations & Offloadable & Non-offloadable \\
\midrule
          None &     Baseline &  14.1 K &     47.2 K &    43.9 &        24.9 &      61.2 K &            68.8 \\
          Data &     Baseline &  14.1 K &       97.7 &    43.9 &        31.1 &      14.2 K &            75.1 \\
          Data &  Partitioned &    29.1 &       97.7 &    43.9 &        31.1 &         127 &            75.1 \\
   Data + pipe &     Baseline &    87.9 &       98.1 &    43.9 &        24.9 &         186 &            68.8 \\
   Data + pipe &     Improved &    5.82 &       19.5 &    43.9 &        6.23 &        25.4 &            50.2 \\
    \hline
 Data + tensor &     Baseline &     879 &       6.10 &    \textbf{2.75} &        1.95 &         885 &            4.69 \\
 Data + tensor &  Partitioned &    1.82 &       6.10 &    \textbf{2.75} &        1.95 &        7.92 &            4.69 \\
            3d &     Baseline &    5.49 &       \textbf{1.31} &    \textbf{2.75} &       \textbf{0.389} &        6.81 &            \textbf{3.14} \\
            3d &     Improved &   \textbf{0.364} &       \textbf{1.22} &    \textbf{2.75} &       \textbf{0.389} &        \textbf{1.58} &            \textbf{3.14} \\
\bottomrule
\end{tabular}

\end{table}

A breakdown of the memory usage for each configuration is shown in table \ref{table:parallel_methods_memory}. All methods are possible from a memory perspective, assuming a minimal memory fragmentation overhead. However, the baseline approach requires an impractical amount of offloaded memory without pipeline parallelism.  The improved method has the lowest memory footprint of 4.72 GB, which is 17 times less than the memory available in an 80 GB A100.

\begin{table}[t]

\end{table}

\paragraph{Smaller clusters}

Although training scales to nearly 40000 GPUs, it may be difficult to find a cluster of that size, so there is a strong case for training with fewer GPUs over a longer period. There are a variety of viable strategies for smaller clusters, allowing trade-offs between efficiency, memory usage, and the choice of parallelism methods. In table \ref{table:smaller_cluster} we provide some example configurations with high efficiency for the target training times of one and six months. With these time constraints, it is possible to train with clusters of sizes 7400 and 1300 respectively, which makes training available to a marginally wider community. The memory usage is increased when compared with the larger clusters but remains far below what is available. Among the methods considered, ours is the most efficient, although the margin becomes negligible for longer training times. It is also far more flexible, as shown for example in the last two entries. For the six-month training it is the only one able to train without tensor parallelism (with offload). It is also able to train with a much lower batch size, which amounts to an extra efficiency gain as there is an implicit cost to training with a high batch size.

\begin{table}[t]
\caption{Selected training configuration for X$_{160}$ for the specified training times of one and six months}
\label{table:smaller_cluster}
\centering
\small
\begin{tabular}{ccccccccc}
\toprule
   Parallelism &       Method &   $b$ & $n_a$ & $n_\text{gpu}$ & Offloadable & Non-offloadable & Efficiency &   Time \\
\midrule
 Data + tensor &  Partitioned &  2415 &    16 &           7728 &        7.92 &            4.69 &       0.93 &   32 d \\
            3d &     Baseline &  2416 &    16 &          10240 &        10.1 &            3.14 &       0.73 &   31 d \\
            3d &     Improved &  2220 &     4 &           7400 &        7.76 &            12.5 &       0.97 &   32 d \\
            \hline
 Data + tensor &  Partitioned &  1660 &     8 &           1328 &        35.0 &            9.38 &       0.97 &  180 d \\
 Pipe + tensor &     Baseline &  2416 &     8 &           1280 &        47.9 &            6.27 &       0.91 &  199 d \\
            3d &     Improved &   792 &     2 &           1320 &        22.4 &            25.1 &       0.97 &  180 d \\
            \hline
   Data + pipe &     Improved &  1572 &     1 &           1310 &        34.2 &            50.2 &       0.98 &  180 d \\
            3d &     Improved &   102 &    16 &           1360 &        11.8 &            3.14 &       0.91 &  186 d \\
\bottomrule
\end{tabular}
\end{table}

\section{Scaling analysis and practical limits}
\label{sec:scaling_analysis}

We analyze how the memory usage and training time scale with the model size, using the X$_[x]$ model family described in appendix \ref{sec:parameter_buffer_details}. We consider the same three training strategies as before --- baseline, partitioned and improved --- but restrict to the fastest configuration for each. 

\begin{figure}[t]
\centerline{\includegraphics[scale=0.5]{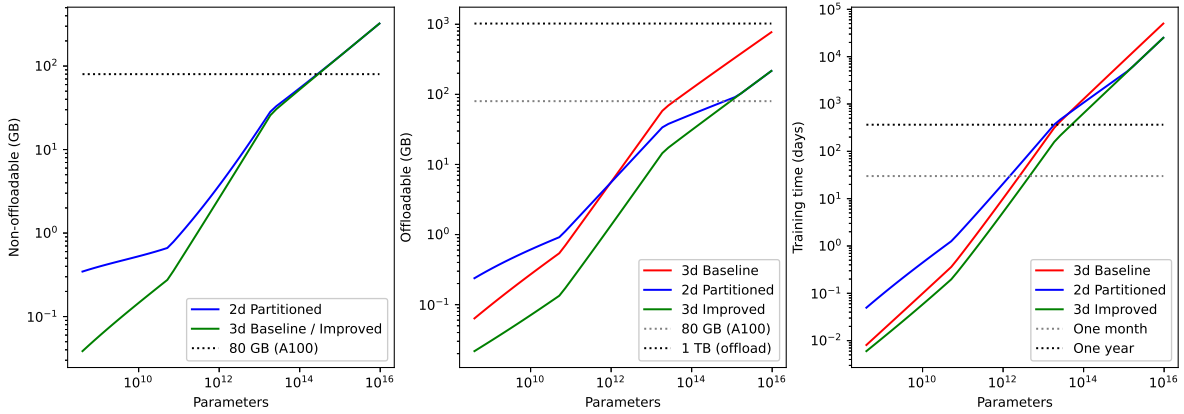}}
\caption{Approximate minimal memory usage and training time for selected training methods, with a maximum node size of 16.}
\label{fig:scaling_limited}
\end{figure}

Figure \ref{fig:scaling_limited} shows the memory usage and training time as a function of the model size at various scales. The improved method outperforms the others at most scales, although it becomes identical to the partitioned approach above the quadrillion parameter scale, as pipeline parallelism is no longer necessary.

Focusing on the improved method, we find that 80 GB is enough to train models up to 280 trillion parameters, with offloading being required above 90 trillion. However, such model would take up to four years to train (and more when taking the real computation efficiency into account), which is unreasonably long as for instance waiting for the next generation of hardware would likely be faster. To obtain better scaling limits, we consider a reasonable threshold of one month, and a more generous threshold of one year. These thresholds respectively result in limits of about 4.5 and 50 trillion parameters, with total memory usages of 13 and 62 GB.

\begin{figure}[t]
\centerline{\includegraphics[scale=0.5]{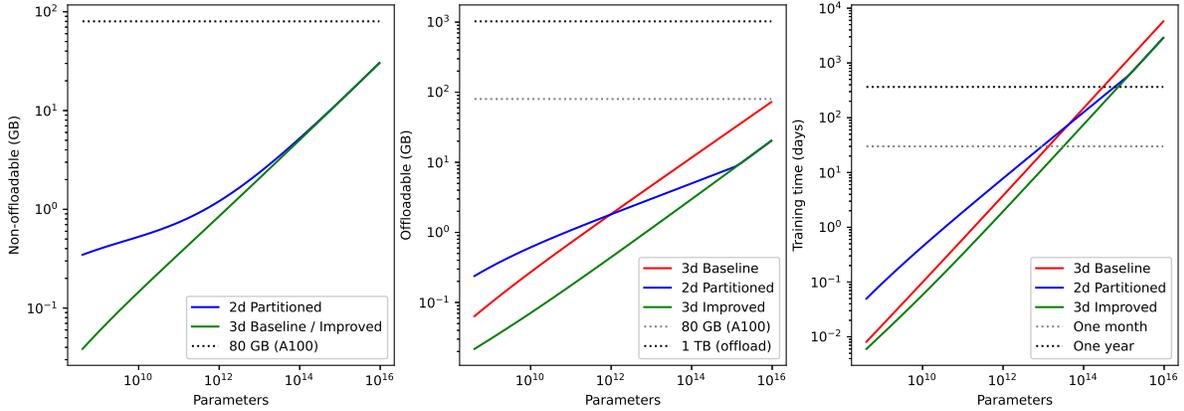}}
\caption{Approximate minimal memory usage and training time for selected training methods.}
\label{fig:scaling_unlimited}
\end{figure}

The above results show a strong limitation on the model size due to the training time. It may however be possible to do better with existing GPUs. While data and pipeline parallelism are fundamentally limited by the critical batch size, tensor parallelism is limited by a computer design choice which limits the node size to 16. The fully connected NVSwitch topology in DGX and HGX nodes is convenient in the general case, but tensor parallelism only requires a ring topology, which is easy to scale to an arbitrary size. This means larger nodes (or separate nodes connected with NVLink) should be possible, although it may require a certain amount of engineering. For this reason, we consider the scenario where the node size limitation is removed, with the results being shown in figure \ref{fig:scaling_unlimited}. In this case, there is enough memory for models up to 100 quadrillion parameters, while the training time reduces the limit to 40 trillion (one month) or 900 trillion parameters (one year).

\paragraph{There is no memory wall}

\begin{figure}[t]
\centerline{\includegraphics[scale=0.45]{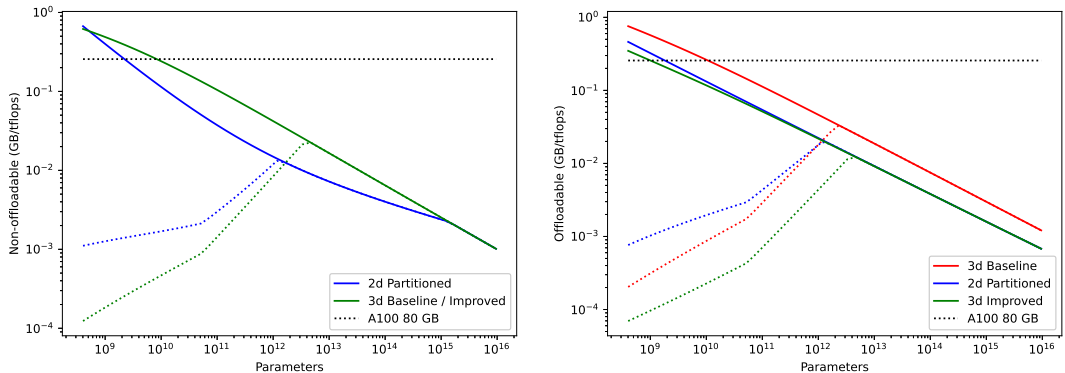}}
\caption{Required memory to compute ratio for training a transformer in one month, as a function of the model size. The dotted line shows a faster training scheme which is already possible on A100s.}
\label{fig:scaling_constant_time}
\end{figure}

While memory usage is not a problem until astronomical scales, we can go further and show memory is \textit{never} a problem. For this, we suppose at some point in the future it becomes possible to train a given model in a fixed time of one month, either using faster devices or faster network connections. Then we measure how much memory would be needed, in relation to the computing power. For simplicity, we assume that the configuration still uses as much data and pipeline parallelism as possible, and scales only with tensor parallelism. The results are shown in figure \ref{fig:scaling_constant_time}, which shows that the memory requirement \textit{decreases} with the model size. In fact, memory is only an issue at smaller scales, which may in part explain the perceived memory problem, but it is already possible to train those models much faster with a minimal amount of memory (dotted line). These results are not particularly surprising, since the highest memory scaling comes from the state which is proportional to the model size, while the computation has a worse scaling due to the increased input size.

\section{Additional training considerations}
\label{sec:additional_considerations}

While we assumed so far the existence of a sufficiently large supercomputer, in practice this may be difficult to achieve. Training a large language model is very expensive, and setting up such supercomputer exclusively for this purpose significantly increases the cost. For this reason, it is preferable to leverage data centers where the nodes can quickly switch to and from other workloads so that they are used without interruption. In this scenario, the number of available nodes may vary during training, for example if higher priority workloads need to be run or some nodes need to be removed for maintenance. This means training should preferably be \textit{elastic}, i.e., it should support variations in the cluster size. Even in a fixed cluster, the risk of hardware failure is high due to the large number of components, and handling these failures requires some form of elasticity. While a complete analysis of elastic training is outside the scope of this paper, in this section we investigate some concerns and potential optimizations related to the optimizations presented in this paper. We assume the nodes are located in the same data center and preferably connected through InfiniBand, although in section \ref{sec:slow_network} we investigate the effect of training over a slower Ethernet connection. We also assume the various parallelism and network bounds presented in this paper are respected, which can be done by scaling the data parallelism degree down from the maximum allowed value. When the state is partitioned, this presents additional challenges since the partitioning changes during training, but we show in section \ref{sec:slow_network} that partitioning actually \textit{helps} with elasticity. The varying partition may also increase the memory usage, but there is plenty of room for it.

\subsection{Don't decay the learning rate, increase the cluster size}

The analysis so far wrongly assumed that the critical batch size is constant during training. During early training the gradients contain a strong model-improving signal, but as the model is trained this signal becomes less important relative to the noise, which increases the number of samples required to obtain an accurate estimate, i.e., the critical batch size. The ``unique'' critical batch size considered so far corresponds to the value during late training, and using it implies a wasteful training above the critical batch size during training. Instead, the critical batch size can be evaluated dynamically during training, and the batch size adjusted accordingly \cite{mccandlish2018empirical,smith2018dont}. In the present context, this means the maximum cluster size varies during training, and when elastic training is possible it should be adjusted dynamically. Doing so reduces the cost of training without significantly affecting the training time.

\subsection{Offloading revisited: real-time checkpoints}
\label{sec:fast_checkpoints}

In the previous sections we found offloading to be possible, but rarely necessary from a memory perspective. Here we take another look at offloading, this time with the objective of keeping a copy of the weights in a secure and accessible location, a practice also known as ``saving a checkpoint''. Saving checkpoints typically takes a long time during which the cluster is idle, often taking longer than training multiple batches. This means checkpoints cannot be saved often, and failures lead to a significant loss of progress. There is also a significant downtime in elastic training, since a variation of the cluster means a checkpoint must be saved, then loaded in the new nodes. This downtime is problematic when running in a data center with a lot of activity, where nodes are added and removed frequently.

\begin{figure}[t]
\centerline{\includegraphics[scale=0.5, trim=0 0 175pt 0]{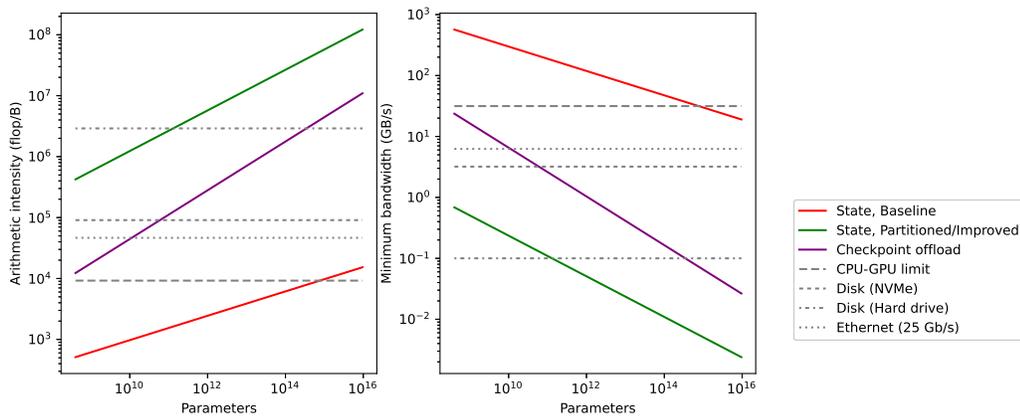}}
\caption{Required memory to compute ratio for training a transformer in one month, as a function of the model size. The dotted line shows a faster training scheme which is already possible on A100s.}
\label{fig:scaling_offload}
\end{figure}

Figure \ref{fig:scaling_offload} shows the arithmetic intensity and bandwidth requirement for offloading the model. In the partitioned case, we find that the state can not only be offloaded to CPU, but it can also easily be offloaded on fast SSDs (as suggested in \cite{rajbhandari2021zeroinfinity}), possibly on a remote location via Ethernet, and for larger models even hard drives are fast enough. This allows keeping an up-to-date copy of the weights in a secure external location accessible by all nodes at a negligible cost, which reduces the potential cost of failure to a single batch\footnote{This can also be done in the non-partitioned by sharing the load of saving the checkpoint across the data-parallel instances, but all nodes still need to load the entire checkpoint.}. The overhead of modifying the cluster size can also be reduced to almost zero by loading the weights on the fly, even if a new partition needs to be made.

We can go further by considering activation checkpoints, which as figure \ref{fig:scaling_offload} shows need a higher bandwidth than the state, but for the larger models can also be saved to a fast remote storage. This reduces the potential loss from a crash to a single layer and allows swapping nodes in the middle of a batch at nearly zero cost.

\subsection{Ethernet is enough}
\label{sec:slow_network}

\begin{figure}[t]
\centerline{\includegraphics[scale=0.5]{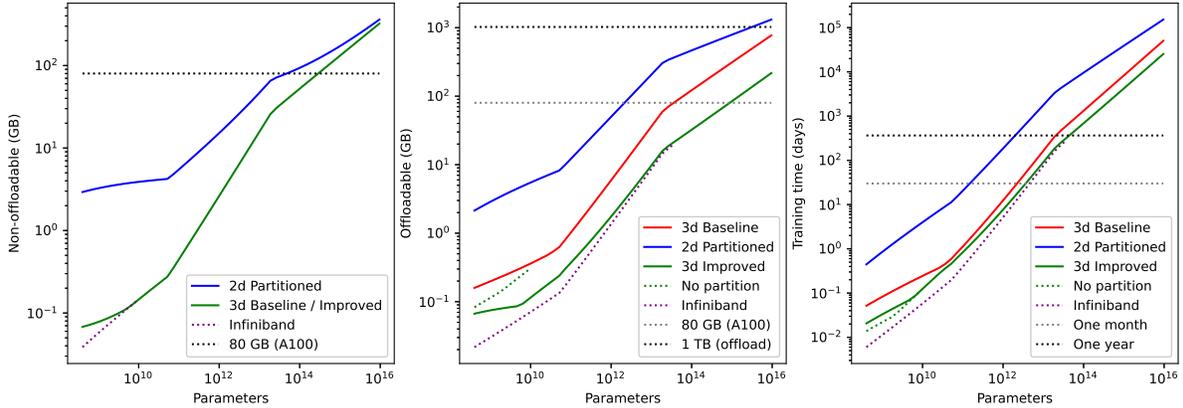}}
\caption{Approximate minimal memory usage and training time for selected training methods, with a 25 Gb/s Ethernet connection.}
\label{fig:scaling_Ethernet}
\end{figure}

As data centers may not be equipped with a fast InfiniBand connection, we evaluate the possibility of training over Ethernet. We assume the nodes are equipped with a 400 Gb/s Ethernet connection, which amounts to 25 Gb/s per GPU. The analysis is shown in figure \ref{fig:scaling_Ethernet}. For larger models, the slower connection makes little difference, provided pipeline parallelism is used. However, for smaller models, a higher degree of pipeline parallelism is required to reduce the network usage, which makes it harder to mitigate the bubble in the improved case. For the trillion-parameter model, this slows down training by about 4\%, but the effect is more important at smaller scales. Despite this and the communication overhead from the state partition, the improved method outperforms the baseline at smaller scales because of the improved communication overlap. For the smallest models, the partition can be avoided to further reduce the training time at a minimal memory cost (dotted line).

\section{Conclusions and future work}
\label{sec:conclusion}

We showed that 3d parallelism is necessary for training large language models, although it remains limited by fundamental and practical bounds affecting the training time. With a combination of layered gradient accumulation, modular pipeline parallelism and training state partition, it is possible to nearly achieve these bounds while using only a small fraction of the available memory. Given a sufficiently large cluster, this allow efficiently training models above the trillion-parameter scale in a reasonable time.

Our approach is also not particularly demanding on the (inter-node) network, being able to perform relatively well with only an Ethernet connection. It also helps with the training of smaller models, which benefit from the improved communication overlap. We expect our methods to allow training models such as BERT noticeably faster than with existing state-of-the-art methods, in a matter of minutes. However, further research would be needed to make accurate predictions for models of that size, as there may be an additional overhead due the small size of the computational kernels.

\subsection{Fine-tuning and inference}

The recent interest in extremely large models has been in large part fuelled by their potential to learn new tasks on the fly, potentially eliminating the need for fine-tuning. When fine-tuning is still needed, it requires much less computing power than training from scratch, but the computational challenge remains to some extent. For example, fine-tuning a trillion-parameter model for 1000 steps requires over 2000 GPU-days, which remains too high for many researchers. Fine-tuning can be done on smaller clusters than those considered in this paper, which may bring back the need for offloading the training state. In that regard, our approach improves over existing method such as the ZeRO family. The reduced need for data transfers allows for an easier offload, while also reducing the activation memory and the network requirement. Note that fine-tuning models above the trillion-parameter scale brings back the need for 3d parallelism and large clusters, assuming a pre-trained model is available to begin with.

Inference represents an additional challenge, because it also needs to be optimized for low latency. Data and pipeline parallelism do not help in that regard, and tensor parallelism is limited by the node size. This sets a lower bound of several seconds per trillion parameters, depending on the sequence length and computational precision. Consequently, the largest language models may not be suitable for real-time applications until with current hardware. As with fine-tuning, offloading may also be necessary for inference, which can run with a minimal number of devices.

\subsection{Computational wall}

Beyond the trillion-parameter scale, our results show an increasingly high lower bound on the training time which makes it impossible to train arbitrarily large models. For example, a 50 trillion parameter model takes at least a year to train, while a quadrillion parameter model would need decades. As hardware speed is currently increasing at a much slower pace than the model size (and the computational requirement), this computational wall will not be addressed by hardware speed alone. Instead, a dedicated effort is needed to either reduce the computational requirement or provide more computational power through distributed training improvements.

\paragraph{Sparse models}

There is a significant research effort dedicated to reducing the computational requirement of language models through sparsity, with some promising results. For example, \textit{mixture of experts} methods have been shown to improve the model performance for a fixed computational budget \cite{fedus2021switch,lepikhin2020gshard}, while \textit{sparse matrix multiplications} enable training on much larger sequences \cite{beltagy2020longformer,child2019generating,zaheer2021big}. While we leave a detailed analysis for future work, we expect sparsity to increase the network and memory usage due to the reduced arithmetic intensity, which should not be problematic unless pushed to the extreme.

\paragraph{Hardware implications}

Recent trends in hardware have focused on large GPUs with as much memory as possible. This was motivated in good part by the requirements of smaller models as well as simplicity considerations, as large devices reduce the need for complex parallelism methods and aggressive memory optimizations. However, this focus on size increases costs, and splits the limited memory bandwidth between large numbers of computational cores, causing memory bottlenecks.

The situation is different for larger models, for which simplicity is not an option. In this case 3d parallelism becomes necessary, and the unavoidable memory optimizations bring down the memory usage to a minimum. Instead, the real challenge for large models lies in the cost and duration of training. As emphasised by our results, the key to a faster training is in large scale tensor parallelism, which needs many GPUs to be connected through a high bandwidth, low latency interconnect. The size of the GPUs is not particularly important,
and in fact, (a lot of) smaller devices may be preferable as they are less affected by bandwidth limitations. The GPUs only need a small amount of fast memory, as long as there is a larger amount of external storage for fine-tuning and inference.

\subsection{Scaling concerns}

While we focused on \textit{how} to train a very large transformer, we did not attempt to determine whether one \textit{should} do it, or even can afford it. Training models at the trillion-parameter scale requires thousands or preferably tens of thousands of GPUs, which implies astronomical costs far beyond the means of most researchers. Training at this scale also has a significant environmental impact, both from the electricity usage and the amount of hardware being used.

In addition to the cost, language models are challenging from an ethical standpoint, and the scale of the model does not help on that regard (see for example \cite{brown2020language}). Language models tend to be highly biased and unfair, a behavior learned from the training data. Due to their performance, they can also be used to impersonate real humans, enabling a wide range of unethical applications. While it is impossible to prevent such misuse in the long term, we hope that large language models become more accessible to the research community in the near future, so that mitigating solutions can be found as soon as possible.

\section*{Acknowledgements}

We would like to thank Harm de Vries for providing extensive support while writing the paper, Eric Robert and Simon B\'elanger for supporting the research project, and Nathan Schucher for providing additional feedback.

\bibliographystyle{plain}
\bibliography{bibliography}

\appendix

\section{Hardware details}
\label{sec:hardware}

This appendix summarizes the hardware specifications relevant to this paper. For more details, see the official documentation~\cite{nvidia}

For the purpose of this paper, the computation is assumed to be done on NVIDIA A100 80 GB GPUs. Such device has a theoretical peak fp16 compute $c_\text{gpu}=312$ Tflop/s, although this value is difficult to reach in practice. The device has $m_\text{gpu}=80$ GB of memory, with a peak bandwidth $\beta_\text{gpu}=2039$ GB/s. Each GPU is connected to the rest of the computer with a PCIe 4.0 x16 connector and can be connected to oher GPUs with up to 12 NVLink interconnects, for a total bandwidth of 300 GB/s in each direction. When combined with NVSwitch, NVLink allows up to 16 GPUs to be fully connected, each pairwise connection being able to fully utilize the NVLink bandwidth.

Although we do not assume a specific computer architecture, we use a 16-GPU DGX or HGX A100 is as a reference\footnote{As no detailed architecture could be found, we extrapolated some details from other DGX and HGX nodes. At the time of writing this paper, there does not appear to be a 16 GPU version for the DGX A100.}. The GPUs are fully connected through NVSwitch, and each pair is connected to a PCI-express switch which also connects to a pair of 200 Gb/s InfiniBand interconnects and to one of the CPUs. While there is a total of 16 InfiniBand connectors, each GPU can only efficiently use one connector due to the PCI-express bottleneck. The CPUs are bundled with a large amount of memory, an Ethernet connection, additional optional InfiniBand connectors, and other standard computer components. The Ethernet connection scales to at least 400 Gb/s, which amounts to 25 Gb/s per GPU\footnote{In a data parallel scenario, network operations use a ring topology which effectively makes the entire Ethernet bandwidth available to a single device. We ignore this fact for simplicity, and in any case, this does not happen with tensor parallelism where all devices communicate with other nodes.}.

The PCI-express configuration of the HGX server creates a bottleneck on the CPU side, since a single PCIe 4.0 x16 connects the CPU to two GPUs and two InfiniBand interconnects, effectively dividing the maximum CPU-GPU throughput by half and preventing the CPUs to efficiently connect with InfiniBand. This is problematic in the offloading scenarios where data is transferred through the CPUs. The problem could be fixed with improved computer designs but remains an important practical limitation.

The various bandwidths and arithmetic intensities are summarized in table \ref{table:hardware_bandwidth}.

\begin{table}[t]
\centering

\caption{Bandwidth and arithmetic intensity with respect to a A100 GPU for the available network interconnects and data storages.}
\label{table:hardware_bandwidth}

\begin{tabular}{ cccccccc } 
\toprule
 \multirow{3}{*}{Network} & Bandwidth & Arithmetic intensity threshold\\
 & Input + Output & @ 312 Tflop/s \\
 & GB/s & flops/B \\
\midrule
GPU memory & 2039 & 143 \\
NVLINK & 600 & 484 \\
PCI-express & 63 & 4.61 k \\
InfiniBand (200 Gb/s) & 50 & 5.81 k \\
CPU-GPU & 31.5 & 9.22 k \\
Ethernet (25 Gb/s) & 6.25 & 46.5 k \\
Disk (NVMe) & 3.2 & 90.8 k \\
Disk (Hard drive) & 0.1 & 2.91 M \\
\bottomrule
\end{tabular}
\end{table}

\section{Transformer scaling}
\label{sec:transformer_scaling}

Transformers are largely insensitive to the exact values of the hyperparameters~\cite{kaplan2020scaling}, assuming they fall within a reasonable range, so the driving factor when scaling the model is computational efficiency. The intermediate size factor $n_I$ should be kept constant, and we use the common value $n_I=4$. The sequence length and head size should be proportional to keep the intermediate activations balanced. The number of heads can scale independently but has an impact on parallelism and the allowed values of the sequence length. In this paper we assume a mild scaling $d_s \sim d_m^{1/2}$, which we achieve with the reasonable relative scaling $d_s = 8d_h = 32 d_a$. These ratios lead to values comparable to what is found in the literature, and match BERT for $d_a=16$, but differ slightly from GPT-3 which gives more weight to the head count.

The layer count scaling is relevant to parallelism, as deeper networks are suitable to pipeline parallelism, while wider ones enable more tensor parallelism. The impact is minimal given that both are limited by other factors, but wider networks are slightly preferable as they lead to more efficient tensor parallelism. In terms of memory, deep and thin networks reduce the size of the buffers and activations, at the cost of larger activation checkpoints. As parallelism is much more important than memory, we select a mild scaling $d_l=\sqrt{d_m}$.

The resulting model family X$_{[x]}$ is parametrized by a single variable $x$:
\begin{align}
&  d_a=\frac12x,
&& d_h=2x,
&& d_l=x, \nonumber\\
&  d_s=16x,
&& d_m=x^2,
&& d_I=4x^2.
\end{align}
Table \ref{table:x_model} shows some examples of X$_{[x]}$ models and a comparison to other large language models.

In~\cite{kaplan2020scaling} it was found empirically that the critical batch size scales approximately as $p^{1/3}$, when measured in tokens. To obtain a numerical value, we assume that GPT-3 was trained at the critical batch size (3.2 M tokens). This results in the empirical formula
\begin{equation}
\label{eq:critical_batch}
b_c \approx 573 \frac{p^{1/3}}{d_s} \approx 82.0 x^{2/3}.
\end{equation}
Although equation \ref{eq:critical_batch} is approximate and was not demonstrated to scale for the whole parameter range studied in this paper, we take it as the true value of the critical batch size for numerical estimations\footnote{The dependence on the sequence length was not demonstrated in \cite{kaplan2020scaling}, and we make the reasonable approximation that the critical batch size measured in tokens does not depend on the sequence length.}.

\begin{table}[t]
\renewcommand{\arraystretch}{1.25}
\centering

\caption{X$_{[x]}$ model configuration examples for a wide range of scales and comparison to some existing large transformer. Parameter counts exclude the embedding layer and language model head.}
\label{table:x_model}

\begin{tabular}{ cccccccc } 
\toprule
Model & $p$ & $b_c$ ($b$) & $d_s$ & $d_a$ & $d_h$ & $d_m$ & $d_l$\\
\midrule
X$_2$ & 488 & 130 & 32 & 1 & 4 & 4 & 2 \\
BERT & 301 M & 751 (256) & 512 & 16 & 64 & 1024 & 24 \\
X$_{32}$ & 403 M & 826 & 512 & 16 & 64 & 1024 & 32 \\
Megatron-LM & 8.15 B & 1130 (512) & 1024 & 32 & 96 & 3072 & 72 \\
X$_{64}$ & 12.9 B & 1310 & 1024 & 32 & 128 & 4096 & 64 \\
T-NLG & 17.0 B & 1440 (512) & 1024 & 28 & 152 & 4256 & 78 \\
GPT-3 & 174 B & 1560 & 2048 & 96 & 128 & 12288 & 96 \\
X$_{108}$ & 176 B & 1860 & 1728 & 54 & 216 & 11664 & 108 \\
X$_{160}$ & 1.26 T & 2420 & 2560 & 80 & 320 & 25600 & 160 \\
X$_{[x]}$ & $12x^5+13x^3$ & $82.0x^{2/3}$ & $16x$ & $\tfrac12x$ & $2x$ & $x^2$ & $x$ \\
\bottomrule

\end{tabular}
\end{table}

\section{Resource usage}
\label{sec:resource_usage}

\subsection{Computation}

In a transformer, as with nearly all deep learning models, the bulk of the computation is performed in the matrix multiplications. These appear in the weight multiplications in the dense layers, and in the self-attention mechanism, but the self-attention matrix multiplications are much smaller in general and can be neglected. For the forward pass, this leads to a computational cost of two floating point operations for each input token and parameter, or $2b d_s p$ flops per batch. In the backward pass, the parameter and layer gradient computation each require a similar amount of computation, to which is added the activation re-computation, for a total of three times the forward pass computation. Summing up, each batch requires $8b d_s p$ flops of computation, or $\tfrac{8b d_s p}{n_\text{gpu}}$ for each device.

\subsection{Parameter and gradient buffering}
\label{sec:parameter_buffer_details}

To determine the memory usage, we need to determine the size and lifetime of the parameter and gradient buffers. Each parameter is used in both the forward and backward pass, so should be restored at least twice. A convenient choice is to define a buffer for all the parameters in a layer, i.e., between two activation checkpoints, and similarly for the gradients. To allow overlapping the communication, two parameter buffers are needed, but one gradient buffer is sufficient. This \textit{mixed buffering} method (as opposed to a strict single or double buffering) is summarized in table \ref{table:mixed_buffering}.

\begin{table}[t]
\centering

\caption{Operation sequences and their resource usage for layer buffering methods. All values are relative to the double buffered forward pass.}
\label{table:mixed_buffering}

\begin{tabular}{ ccccccc } 
\toprule
Stream 1 & Stream 2 & Parameter & Gradient & Computation & Network & Arithmetic  \\ 
 & & buffers & buffers & & & Intensity \\ 
\midrule

\multicolumn{7}{c}{Forward} \\
Activations$(i-1)$ & Restore$(i)$ & 2 & 0 & 1 & 1 & 1\\ 
Activations$(i)$ & Restore$(i+1)$ & 2 & 0 & 1 & 1 & 1 \\ 

\multicolumn{7}{c}{Backward} \\
Gradients$(i-1)$ & Restore$(i)$ & 2 & 1 & 2 & 1 & 2 \\ 
Activations$(i)$ & Reduce$(i-1)$ & 1 & 1 & 1 & 1 & 1 \\ 
Gradients$(i)$ & Restore$(i+1)$ & 2 & 1 & 2 & 1 & 2 \\ 
Activations$(i+1)$ & Reduce$(i)$ & 1 & 1 & 1 & 1 & 1 \\ 
\bottomrule

\end{tabular}
\end{table}

Note that in the absence of gradient accumulation, it is possible to achieve a lower memory usage with buffers defined at the sub-layer level, possibly even splitting the parameter within a single operation as suggested in \cite{rajbhandari2021zeroinfinity}. Doing so requires restoring the parameters an extra time in the backward pass, but this can be done without increasing the network requirement by leveraging the arithmetic intensity imbalance (see table \ref{table:mixed_buffering}). However, with gradient accumulation, the network operations need to happen for each micro-batch, which leads to an excessive network requirement and defeats the purpose of layered gradient accumulation. In any case, we find mixed buffering sufficient for all realistic scenarios.

\subsection{Memory}

As described in section \ref{sec:memory_optimizations}, the memory usage breaks down into four categories: the training state, the activation checkpoints, the parameter and gradient buffers, and the layer activations. The first two can be offloaded to CPU memory, but not the latter two which ultimately determine how big the model can grow from a memory perspective.

With the Adam optimiser, the training state consists of the model parameters as well as their running mean and variance, all stored with single precision, for a total of $12 p$ bytes. The gradients would take an extra $2 p$ bytes, but we can e this to a negligible amount by updating the weights as soon as possible. In the non-partitioned case, the state is split across the model-parallel instances ($\tfrac{12 p}{n_l n_a}$ bytes per device), while in the partitioned case it is split across all devices ($\tfrac{12 p}{n_\text{gpu}}$ bytes each).

In the mixed buffering method, two parameter and one gradient buffers are needed, each being the size of a single transformer layer. The buffers are split in the tensor parallel dimension for a total of $\tfrac{6 p_l}{n_a}$ bytes per GPU. Note that while the buffers are much smaller than the training state, they do not shrink much with parallelism so can still be important.

The checkpoints are assumed to match the output of each transformer, which works relatively well in practice, for a total of $2 b d_s d_m d_l$. They split naturally over the data and pipeline parallel dimensions and can also be partitioned in the tensor parallel dimension, for a memory usage of $\tfrac{2 b d_s d_m d_l}{n_\text{gpu}}$ bytes per device.

This formula is not optimal, as for example some intermediate computations can be combined together into fused kernels, but it is sufficient for the purpose of this paper as the activation memory remains low enough. The activation memory is split between the micro-batches and the tensor-parallel instances, for a total of $\tfrac{b d_s m_0}{n_b n_\mu n_a}$

\subsection{Network}

3d parallelism involves three kinds of network communication, one for each parallelism dimension. This section aims at evaluating the bandwidth requirement for each, as well as the associated arithmetic intensity. For the communication to be overlapped perfectly, the arithmetic intensity $\nu_{op}$ for the computation with respect to the network transfer needs to be higher than the arithmetic intensity $\nu_{net}$ implied by the GPU and the network, or
\begin{equation}
\label{eq:overlap_condition}
    \nu_{op}\geq\nu_{net},
\end{equation}
When efficient overlap is not possible, there is a relative overhead $\tfrac{\nu_{net}}{\nu_{op}}$. This overhead is expected to remain within a chosen threshold $\epsilon$, resulting in the condition
\begin{equation}
\label{eq:overhead_condition}
    \epsilon~\nu_{op}\geq\nu_{net}.
\end{equation}

\subsubsection{Data parallel}

In the non-partitioned case, the gradient reduction involves a scatter-reduce and an all-gather, which are both identical from a computational perspective and are generally implemented with bandwidth-optimal ring methods. In an all-gather for instance, each device receives all the gradients except for the ones it already has and sends the same amount of data. For the gradient reduction, this results in a bandwidth usage of $\tfrac{8(n_b-1)p}{n_\text{gpu}}\approx \tfrac{8p}{n_l n_a}$. The reduction is overlapped with the backward pass for the last micro-batch, except for boundary effects at the first and layers. When comparing with the backward pass compute of $\tfrac{6b d_s p}{n_\mu n_b n_l n_a}$ resulting in an arithmetic intensity
\begin{equation}
\label{eq:data_intensity_base}
    \nu_b^{\text{base}}\approx\frac{3b d_s}{4 n_b n_\mu},
\end{equation}
assuming $d_n/n_l$ is large enough. This means the overlap gets worse with data parallelism and micro-batching, and by extension with pipeline parallelism. In the latter case case this is due to poor communication overlap, and for $d_l=n_l$ there is no overlap at all. Because of that, the non-overlapped scenario is more appropriate, with
\begin{equation}
\label{eq:data_intensity_base_pipe}
    \nu_b^{\text{pipe}}\approx\frac{b d_s}{n_b}.
\end{equation}

In the partitioned case, there is an extra all-gather in the forward pass, and the network operations need to be done for each micro-batch, resulting in $\tfrac32n_\mu$ times the network bandwidth when compared with the non-partitioned case. The lowest arithmetic intensity is in the forward pass, with the value
\begin{equation}
\label{eq:data_intensity_base_partition}
    \nu_b^{\text{base-part}}\approx\frac{b d_s}{2 n_b n_\mu}.
\end{equation}
This is only 33\% lower than without the partition, but in the micro-batched case the overlap is with all the micro-batches rather than the last one, so the non-overlapped scenario does not apply.

With layered gradient accumulation, the network usage remains the same, but can be overlapped with the entire backward pass even in the micro-batched case (again excluding boundary effects). This remains true with pipeline parallelism, in the recommended setting where $\tfrac{d_l}{n_l}$ is not too small. This results in the arithmetic intensity
\begin{equation}
\label{eq:data_intensity_improved}
    \nu_b^{\text{impr}}\approx\frac{3b d_s}{4 n_b},
\end{equation}
or
\begin{equation}
\label{eq:data_intensity_improved_partition}
    \nu_b^{\text{impr-part}}\approx\frac{b d_s}{2 n_b},
\end{equation}
depending on whether the state is partitioned.

\subsubsection{Pipeline parallel}

With pipeline parallelism, each instance needs to receive its inputs and send its outputs, and the potential bottleneck is in the forward pass. In the baseline, each micro-batch the forward pass involves $\tfrac{4 b d_s d_m}{n_b n_a}$ bytes of communication and $\tfrac{2 b d_s p}{n_\text{gpu}}$ flops of computation, for an arithmetic intensity of
\begin{equation}
\label{eq:pipeline_intensity_base}
    \nu_l^{\text{base}}\approx\frac{(2+n_I)d_m d_l}{n_l}.
\end{equation}
With modular pipeline parallelism, the number of transfers is multiplied by $\tfrac{d_l}{n_l}$, for an intensity
\begin{equation}
\label{eq:pipeline_intensity_impr}
    \nu_l^{\text{impr}}\approx(2+n_I)d_m,
\end{equation}
which is high enough for large models.

The data transfer can be overlapped with the computation, but it is difficult to do so with the minimal number of micro-batches $n_{\mu}=n_l$. This can be addressed by adding a small number of extra micro-batches, approximately given by $\tfrac{\nu_l}{\nu_{net}} n_\mu$, or more is the network speed fluctuates.

\subsubsection{Tensor parallel}

Following the approach of \cite{shoeybi2020megatronlm}, tensor parallelism requires two all-reduce operations for each transformer layer in the forward pass, which are not overlapped with computation. An extra two all-reduces are needed in the gradient computation, for a total of six (when including the activation re-computation). For a given layer, this implies a network usage of $\tfrac{24 b d_s d_m (n_a-1)}{n_b n_a}$ bytes compared with a compute of $\tfrac{8 b d_s p_l}{n_b n_a}$ flop, for an intensity
\begin{equation}
\label{eq:tensor_intensity}
    \nu_a\approx \frac{(4+2 n_I)d_m}{3 (n_a-1)}.
\end{equation}

\subsection{CPU-GPU transfers}

Offloading the training state and activation checkpoints requires large data transfers between the CPU and GPU memory. For the state offloading, the computation for a given layer is overlapped with the transfer of the parameters and the gradients for a given layer or its partition. In the baseline, the transfer needs to happen for each micro-batch, while with layered gradient accumulation it happens once for all micro-batches. The bottleneck is in the forward pass, with four possible values depending on the scenario:
\begin{align}
\label{eq:state_offload_intensity}
 & \nu_s^{base}\approx \frac{b d_s}{n_\mu n_b},
 & & \nu_s^{base-part}\approx \frac{b d_s}{n_\mu}
 & & \nu_s^{impr}\approx \frac{b d_s}{n_b},
 & & \nu_s^{impr-part}\approx b d_s
\end{align}
For checkpoint offload the computation is similar to the pipeline-parallel network transfer, with half the computation, for an intensity
\begin{align}
\label{eq:checkpoint_offload_intensity}
\nu_c\approx(4+2n_I)d_m
\end{align}

\end{document}